# Annotating High-Level Structures of Short Stories and Personal Anecdotes

Boyang Li[1], Beth Cardier[2], Tong Wang[3], and Florian Metze[4]

[1] Independent
[2] Sirius-Beta, Virginia Beach, VA
[3] University of Massachusetts Boston, Boston, MA
[4] Carnegie Mellon University, Pittsburgh, PA
boyangli@outlook.com, bethcardier@sirius-beta.com, tong.wang001@umb.edu, fmetze@cs.cmu.edu

**Abstract**

Stories are a vital form of communication in human culture; they are employed daily to persuade, to elicit sympathy, or to convey a message. Computational understanding of human narratives, especially high-level narrative structures, remain limited to date. Multiple literary theories for narrative structures exist, but operationalization of the theories has remained a challenge. We developed an annotation scheme by consolidating and extending existing narratological theories, including Labov and Waletsky's (1967) functional categorization scheme and Freytag's (1863) pyramid of dramatic tension, and present 360 annotated short stories collected from online sources. In the future, this research will support an approach that enables systems to intelligently sustain complex communications with humans.

## 1. Introduction

Story is a fundamental form of human communication, sometimes argued to be more powerful than logical arguments (Bruner 1986; Fisher 1987). Stories can be used, for example, to persuade, to encourage, to elicit sympathy, and convey a moral, message, value or lesson. It follows that a computational understanding of stories will help computer systems communicate better with users. Recent years have witnessed growing interests in computational approaches for story understanding (Bamman *et al.* 2013; Ferraro and Van Durme 2016; Finlayson, M. A. 2016; Goyal *et al.* 2010; Ouyang and McKeown 2015; Huang *et al.* 2016; Pichotta and Mooney 2016; Tapaswi *et al.* 2016; Mostafazadeh *et al.* 2016; Chaturvedi *et al.* 2017; Wang *et al.* 2017; Dogan *et al.* 2018). Yet few attempts to understand high-level story structures, which are the focus of the present paper.

What constitutes story structure or a "story arc" may be debatable since there is more than one facet to a story. As an operating definition, we consider *story structure* to satisfy the following requirements: (1) it contains a small set of functions with typical orderings between them, though atypical orderings are sometimes possible. (2) The functions are independent of content and genre; they describe structures of stories with different content in any genre. (3) The functions carry significance on the dramatic arc, and (4) together they describe most of a story rather than a small part of it.

This definition rules out the functions proposed by Propp (1928) for Russian folklores, components of the hero's journey (Campbell 1949), and other similar theories because they are closely tied to one type or genre of stories and are not domain-independent. Event-level representations, such as plot units (Lehnert 1981), also do not fit the definition because not all events play important dramatic roles and the ordering between events can be rather arbitrary.

Instead, we investigate what was described by Aristotle (circa. 335 BC) as the beginning, the middle and the end of a story. Similar to Aristotle, Freytag (1863) proposed a dramatic structure containing five parts, whose modern version includes Exposition, Rising Action, Climax,

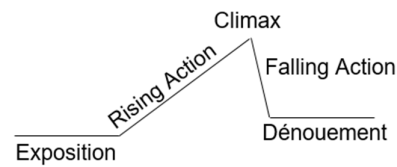

Figure 1. Freytag's story structural functions. The line indicates how dramatic tension heightens and lowers throughout the story.

Falling Action, and Dénouement. The parts are correlated with the rising and falling of dramatic tension, as illustrated in Figure 1. Labov and Waletzky's theory on narrative analysis (1967; Labov 2013) (henceforth L&W) provides another structure that starts with *Abstract* and *Orientation*, goes through *Complicating Actions*, *The Most Reportable Event*, *Evaluation*, to end with *Resolution* and *Coda*. In our opinion, structural functions proposed by the two theories both satisfy the four requirements laid out earlier.

Although these theories seem reasonable by themselves, some important open questions remain: (1) Are these theories compatible or mutually exclusive? If they are compatible, do they describe the same story stages using different terms? (2) Can we devise a unified version of these theories and operationalize it sufficiently, so that human annotators can reliably annotate stories using the theory?

In this paper, we attempt to answer these questions. First, we identify similar concepts that are described by both theories. Based on this understanding, we develop a new annotation scheme, which reconciles the two theories and provides additional functions that we find useful in annotating casual stories online. Finally, we trained annotators to label sentences in stories acquired from online sources and public datasets, yielding 360 unique annotated stories.

To our knowledge, this is the first effort aimed at creating an operational annotation schema that unifies different accounts of story macro-structures. Previous annotation schemata either focus on event-level representations that do not always have dramatic significance (Elson 2012;

| Freytag | L&W | Prince | Todorov | Our Annotation |
|---|---|---|---|---|
| Exposition | Orientation | Starting State | Old Equilibrium | Orientation |
| Rising Action | Complicating Actions | | Disruption | Complicating Actions |
| Climax | Most Reportable Event | State-changing Event | Efforts to repair the disruption | Most Reportable Event |
| Falling Action | Resolution | Ending State | | (Minor) Resolution |
| Dénouement | Coda | | New Equilibrium | Aftermath |

Table 1. Correspondence between categories from different narrative theories and our annotation.

Lehnert 1981), or they use only part of the dramatic curve (Ouyang and McKeown 2015).

## 2. Related Work

There have been several attempts at annotating story semantics and computationally predicting semantic labels from text. Ouyang and McKeown (2015) (henceforth O&M) identified the most reportable event (MRE) in L&W as the "nucleus" of the story. Consequently, they annotated the MRE in roughly 500 stories collected from Reddit and built a classifier for identifying MREs from text, but omitted other categories from the theory. Rahimtoroghi *et al.* (2013) and Swanson *et al.* (2014) used a subset of categories from L&W, including orientation, action, and evaluation.

At the event level, Elson (2012) designed an annotation schema, Story Intention Graph (SIG), that captures timelines as well as beliefs, intentions and plans of story characters. We perceive similarities between this annotation and approaches for generating stories and character behaviors, such as Belief-Desire-Intention agents (Rao and Georgeff 1995) and intention-based story planning (Riedl and Young 2010). Lukin *et al.* (2016) annotated 108 personal stories using the SIG formalism. Finlayson (2016) produced extensive annotations for Propp's Russian folklores, ranging from co-reference and temporal ordering to semantic roles and word senses. Gervás *et al.* (2016) found Propp's functions limited to a single genre and created a new set of functions for annotating 42 musicals. In contrast to L&W's theory, we consider these annotations to be on the micro-structure of events rather than the macro-structure of the entire narrative.

Another influential event-based schema of story structure is plot units (Lehnert 1981). In this schema, an event is classified according to its sentiment as positive, negative, or a mental state with neutral sentiment. In addition, the schema further contains four types of causal links between the events: motivation, actualization, termination, and equivalence. These entities form basic plot units. For example, the pattern *success* contains an actualization link going from a mental state to a positive state. Lehnert further argued these basic units can be combined further to eventually capture story-level structure. Appling and Rield (2009) and Goyal *et al.* (2010) trained machine learning models to predict plot unit structures.

A bottom-up approach employs statistics from local regions of text to represent story structure, instead of a predefined set of function labels. Elsner (2015) collected frequency trajectories of character names combined with words expressing emotions and appearing in Latent Dirichlet Allocation (Blei *et al.* 2003) topics to represent story structure and measured similarity between trajectories. Reagan *et al.* (2016) extracted sentiment (i.e., the positive and negative polarities of emotions) from 1,327 story texts and identified 6 common patterns of how sentiment change throughout the stories.

To situate ourselves with regards to previous work, this work adopts the macro-structural view of L&W and Freytag, which differs from the event-based view or the bottom-up view of narrative structure. We propose a new functional schema that reconciles Freytag's theory with L&W, which we believe captures both dramatic tension and the social aspect of online narratives.

## 3. The Annotation Schema

In this section, we start by discussing several narrative theories, including Freytag (1863), L&W (1967; Labov 2013, 1997), Prince (1973), Todorov (1971), and O&M (2015). Table 1 shows the correspondence we identified between different narrative theories. After that, we propose a new set of functional labels that unify fundamental ideas from these theories.

### 3.1 Integrating Narrative Theories

Freytag's five-stage theory of story development in theater (1863) mainly follows the building and resolution of dramatic tension. Modern interpretations of the theory (e.g., Thursby 2006) generalize stories across numerous genres and media. The first stage, *Exposition*, introduces the narrative setting and has the lowest tension. Tension then increases during a process referred to as *Rising Action*, propelled by a crisis. Freytag's tension peaks at the *Climax*, where the forces of tension are concentrated. After the climax, we draw from Thursby's (2006) modern interpretation, where tension quickly falls towards a *Resolution* and then *Dénouement*.[1] In both professional and everyday instances of modern narratives, we have observed a swift drop in tension during this last stage, its fast resolution standing in contrast to the labor by which tension was built.

This pyramid structure is reminiscent of Todorov's analysis (1971) in which a story starts with an equilibrium, which is later disrupted. Efforts to restore the equilibrium are made and the new equilibrium is created in the end. In our interpretation, an equilibrium state has low tension. The disruption leads to high tension and the restoration of equilibrium lowers the tension. Therefore, we align the old equilibrium with Exposition, the disruption with the Rising Action, and the new Equilibrium with the Dénouement.

---
[1] In its second half, Freytag's original framework closely follows the nuances of tragic theatre, featuring several complicated turns of action that are difficult to generalize to other genres.

In comparison, L&W's story structure focuses on the social relationship between the storyteller and the audience and on the surface shares little with Freytag and Todorov. The theory contains the following categories: Abstract, Orientation, Complicating Actions, The Most Reportable Event, Evaluation, Resolution and Coda. Labov argued that the entire story's purpose is to serve the MRE, which is "the event that is less common than any other in the narrative and has the greatest effect upon the needs and desires of the participants in the narrative (is evaluated most strongly)" (Labov 1997, p. 406).

Ouyang and McKeown (2015) went a step further by merging L&W with Prince's (1973) three basic states: the starting state, the ending state, and the transformational event in the middle. Hence, they provided a slightly modified definition for MRE as "the most unusual event that has the greatest emotional impact on the narrator and the audience". O&M further note the Orientation is the starting state and the Resolution is the ending state.

However, we have not yet found a correspondence for Rising and Falling Actions in L&W's framework. In Labov's scheme (1997, 2013) the Complicating Action is any event in a causal sequence, of which the MRE is one. Here we apply an additional requirement that a complicating action must cause the tension to rise, and must make the MRE causally possible. That is, it must cause something to become more complicated, as the name implies.

We also deviate from O&M by aligning Falling Action with Resolution and Dénouement with Coda. Labov defines Coda in terms of its ability to resolve all further questions the audience may have, "so that the question: 'What happened then?' is no longer appropriate" (Labov 1997, p. 402). We understand this as that a new equilibrium has been established and dramatic tension is minimum.

L&W's Abstract and Evaluation do not have corresponding functions in Freytag and others. We attribute this to difference in medium and context – L&W focused on oral stories, which are usually short and less formal than professional productions; the relationship between the storyteller and the listener is usually close. As such, L&W's Abstract draws attention from the listener and signals the following story. An example is when a friend calls and says, "I just had the most amazing experience at the park!" Evaluation usually provides a personal viewpoint from the storyteller, such as "That's why I avoid that restaurant." This type of message is rare in formal narrative productions, except perhaps for children's stories and fables.

The deliberate nature of the above discussion is to discover commonalities among narratological theories that appear different, at the possible risk of not being meticulously faithful to their authors' intentions. We believe this approach provides important insights, especially for operationalizing the theories into a practical annotation schema, which we present below.

## 3.2 The Proposed Annotation Schema

Based on the above theoretical analysis, we present an operationalized theory in terms of narrative functions that we use to label stories. A key practical consideration is to reduce ambiguity in the definitions, so the schema can be easily communicated and the number of ways that a story may be annotated is reduced. Here we describe the 10 functional labels.

A central idea throughout these 10 categories is the *story frame*. We consider events recounted as part of the story as within the story frame. From time to time, the narrator may step outside the frame to reflect on the tale's meaning or connect with the audience. Labov also observed two modes of engagement – one socially-oriented and another in which the speaker is "reliving events of his past" (Labov, 1972, p. 354).

**Abstract**: An abstract is a summarizing account of the key ideas in the tale, and is almost always found at the beginning of the text. Although it contains information about the story (including the gist of the MRE), it does not introduce the inciting action and thus sits outside the story frame. This label can also apply to a story title.

**Orientation**: This is the starting state of the story and thus, like the other stages that successively follow, it sits within the story frame. The orientation consists of a survey of the elements that set up the central action, which may include "time, place, persons and their activity or situation' (Labov, 1972, p. 364). It may also include general tendencies of a person or situation, such as "my brother is usually very healthy" or "my house is always cold".

**Complicating Action**: In general, a complicating action is a single event that increases the tension of the story. It also causes a situation to turn away from everyday routines and become remarkable. Finally, it has a causal component, in that it propels the critical action of the story towards the MRE. We use this label multiple times to indicate a series of complicating actions that build tension with each occurrence.

**The Most Reportable Event**: This is an event that introduces tension, in the same manner as a complicating action, but it also has some unique qualities that means there can only be one in a tale. A sentence or sentences qualify as an MRE if two criteria are fulfilled: (1) it is an explicit event at the highest tension point of the story. (2) If you only report one event as the summary of the story, it is this one.

**Minor Resolution**: This is an explicit event that allows tension to drop slightly during a series of complicating actions. It can occur in two ways: (1) by resolving a lesser mystery in a story, or part of it; (2) by resolving the tension of part of a problem in the story, without resolving the issues of the entire narrative.

**Return of MRE**: If the MRE theme comes back later after the resolution in a new way, either in time or in action, we say it is a 'Return of MRE'. This event is a new twist on the main theme. It must be at similar level of tension and importance as the MRE; it is also separated from the MRE by time or other narrative functions (if not, it is simply the same event as the MRE). On the tension curve, the Return of MRE allows the tension to rise again after the Resolution.

| Story Text | Annotation |
|---|---|
| Yes. A hedgehog saved my life when I was at college. | Abstract |
| I used to work nights and cycle home at 4 am. There was a long decline, then a bend in the road, then an incline. I would usually cycle around the bend full speed, and rely on my hearing and the glare of headlights to see if anyone was coming. | Orientation |
| One night I was cycling home with headphones in as I'd had a bad shift, and going along full speed I saw a hedgehog in the road, and braked just before the bend. Usually I would have just dodged the hedgehog, but the song I was listening to just made me feel like stopping and watching it for a second. | Complicating Action |
| Just as I pulled up, a truck came around that bend at that moment full speed, taking the racing line with its headlights off. I would have not have [sic] even felt a thing. | MRE |
| The hedgehog was killed. | Resolution |

Table 2. An annotated story example with a canonical story structure from Abstract to Resolution.

**Resolution**: This event on the main causal chain happens after the MRE and resolves the dramatic tension of the story. Hence, it is often a concluding action of the story, but can be followed by the Aftermath or the Evaluation.

**Aftermath**: This event occurs when a significant temporal gap has elapsed after the main event sequence has concluded. It indicates the long-term effect or broader implications of the recounted events – for example, how the story characters went on with their lives after the main events are over.

**Evaluation**: This is a comment from the narrator about the significance or meaning of the story itself and is focused on a moral, message, value or lesson. It could even be the absence of a lesson, such as "I didn't learn X". The storyteller stops recounting events to the audience and "tell[s] him what the point is" (Labov, 1972, p. 374). It aligns with Labov's notion of "external evaluation" (Labov, 1972, p. 371). This kind of comment occurs outside the story frame and is usually located after resolution or aftermath.

**Direct comment to audience**: A direct comment openly addresses the audience outside the story frame, for example: "You're not going to believe this." It can also include the reason for telling the story, an apology for the way the story is presented, or concern that telling the story will get the writer into trouble.

### 3.3 Annotated Examples

Tables 2 and 3 show two story examples annotated by the authors. The two stories were collected from Quora, as described by Wang *et al.* (2017) and contain small spelling and grammatical errors common to online, non-professional contributors.

| Story Text | Annotation |
|---|---|
| Spanish captain was walking on his ship. | Orientation |
| A soldier rushes to him and says, "An enemy ship is approaching us". | Complicating Action 1 |
| Captain replies calmly, "Go get my red shirt". | Complicating Action 2 |
| The soldier gets the shirt for the captain. | |
| The enemy ship comes in; heavy rounds of fire are exchanged. Finally, the Spaniards win. | Minor Resolution |
| Soldier asks, "Congrats Sir, but why the red shirt?" | |
| Captain replies, "If I got injured, then my blood shouldn't be seen, as I didn't want my men to loose [sic] hope." | MRE |
| Moral: For success, hope is very important. | Evaluation |
| Just then, another soldier came in and said, "Sir, we just spotted another 20 enemy ships!" The captain calmly replied, "Now Go bring my yellow pants".. :p :p | Return of MRE |

Table 3. An annotated story example showing Evaluation and Return of MRE.

The first story has a canonical structure from Abstract to Resolution. The Orientation provides the background of the story and the storyteller's general tendency to cycle too fast on a curvy road. Strictly speaking, the first half of the next sentence "one night I was cycling home …" can be seen as part of Orientation. However, for consistency we do not allow one sentence to be broken into multiple parts with different annotation labels. The story annotator must decide the focus of the sentence, which is the action of braking. This action is a Complicating Action as it increases tension and is on a causal chain leading to the MRE. The event of the hedgehog being killed is considered Resolution rather than Aftermath because it happens immediately after the MRE.

The second example is likely a retelling of a popular joke. The story's structure is less complete than the first one. In this example, the tension in the story rises again in the end with the Return of MRE and never gets fully resolved. We note this is a common structure in jokes, presumably to surprise the audience and let them figure out the outcome (Li 2016). This story has two Complicating Actions because they are two separate events and each raises tension. This contrasts with the first story's Complicating Action, which has a lengthy description focusing on the single action of braking. The additional complicating actions indicate the greater structural complexity of the second example, in terms of building and managing tension.

Figure 2. The confusion matrix for narrative functions.

## 4. The Annotation Procedure

The stories being annotated come from three sources: stories collected from Quora by Wang *et al.* (2017), stories collected from Reddit by O&W, and stories annotated by Lukin *et al.* (2016). For the Quora stories, two annotators determined if a text is a story and fits our purpose. The criteria include: the text must contain an MRE and is composed of only one story (not multiple stories); the text is shorter than 700 words, longer than 90 words and has less than 6 lines of dialogue; non-narrative elements, such as lengthy reflections that do not drive the story forward, must be less than 50% of the text. Stories that do not meet these criteria are rejected. Stories that contain offensive content are not annotated.

The annotation was mainly performed by two annotators who did not have backgrounds in linguistics or literature. They were asked to first read the entire story and pinpoint the most reportable event. With the MRE determined, they subsequently identify complicating actions on causal chains leading to the MRE, and then the Resolution, where the tension is resolved. The rest of the categories can be identified against this skeleton. They then broke up the text into sentences, which are the basic unit of annotation, and assign them to categories. One sentence cannot be broken into multiple parts with multiple labels. The annotators were encouraged to think in terms of events rather than raw text.

We adopt the following training and validation process. The annotators went through three rounds of tutorials over five weeks, and during each session they were given 25 stories to annotate. These annotations were then compared to the gold standard provided by the authors and corrections were explained. After training, the two annotators annotated the same 71 stories in order to compute interrater agreement. Subsequently, they annotated separate stories. One author of this paper also annotated a small number of stories.

In total, 480 stories containing 8,908 sentences were annotated. Excluding repeated stories, we obtain 360 unique stories, including 167 from the Quora dataset, 73 from Lukin *et al.* and 120 from O&W. A story contains 18.34 sentences on average.

## 5. Validation and Discussion

We computed interrater agreement using Cohen's Kappa between pairs of annotators separately, as not all annotators worked on the same set of stories. The agreement is computed at the sentence level. Among the two annotators and an author, the pair-wise kappas are 0.39, 0.41, and 0.42 respectively, indicating fair agreement among the annotators.

We further analyze the disagreements made by the annotators. Figure 2 shows a detailed confusion matrix among narrative functions and an additional "unlabeled" category. The numbers in the matrix are computed as follows. If annotator A and annotator B agree that a sentence is in category $i$, the count for the cell $(i,i)$, denoted as $c_{ii}$, is incremented by 1. If one labels the sentence as category $i$ and the other labels it as category $j$, the count for cells $(j,i)$ and $(i,j)$ are both increased by 0.5. Finally, the cells are normalized as $2c_{ij}/(\sum_k c_{ik} + \sum_k c_{kj})$.

From Figure 2, we observe that substantial agreement is achieved around the major categories of story structure that appear in all three schemes (ours, Labov's and that of Freytag/Thursby). These are the Orientation, Complicating Action, and MRE. We attribute this high agreement to the observation that the core progressions in a story's structure are usually less ambiguous than the rest.

The three categories close to the end of the story, Resolution, Evaluation, and Aftermath, tended to be mixed up by annotators. Early elements such as Abstract and Orientation also tended to get confused. Although these categories have clear definitions, the differences between them were finer, and in the wild terrain of real-world anecdotes these differences were harder to reliably identify.

This suggests our annotation schema can differentiate major components of the story structure, even though the annotation gets less accurate on the categories that are more specific to particular nuances of story structure. Infrequent categories such as Return of MRE and Minor Resolution are also difficult to annotate. After merging Resolution, Evaluation, and Aftermath into a single category, and treating Minor Resolution and Return of MRE as unlabeled, the three interrater agreement measures increase to 0.44, 0.49, and 0.47, respectively.

We reckon that being able to differentiate major categories across entire narratives is an achievement, especially given the complexity of the annotation scheme and the potential for ambiguity in real-world, non-professionally written texts.

We further note the high cognitive load created by this annotation task. The annotators need to keep the entire story in mind while evaluating each sentence's role in the entire story. In addition, they often need to mentally parse sentences to separate several events and recognize the most important event being described. In the future, the use of intelligent annotation tools could simplify the task and boost interrater agreement. For example, the annotation tool may decompose the annotation of one story into many smaller tasks and reduce the cognitive load. The annotation tool may also utilize semantic role labeling to highlight different events and help the annotator recognize major events. Reducing the number of categories is also likely to further improve interrater agreement.

## 6. Conclusions

Understanding the macro structures of a narrative, such as where dramatic tension rises and falls, is an important link in enabling computer systems to understand the larger dynamics of narrative communication. Existing work tends to focus on categories that are specific to one genre and types of stories or a subset of the story structure.

In this paper, we provide a first attempt at integrating multiple narratological accounts to capture fundamental and holistic structures of a story. To do this, we propose a set of narrative functions that represent the overlap between schemata proposed by Labov and Waletzky (1967) and Freytag (1863), thereby capturing dramatic tension and the social aspect of story structure. We annotated 360 unique stories from three story sources in the literature and achieved fair interrater agreement on the annotations. Upon close inspection, we note confusion in the annotations are concentrated on a few fine-grained categories whereas the core stations of story progression were consistently identifiable. The annotation results suggest the annotation scheme allows the separation of major structural elements, despite the difficulty of the task. We believe this research will lead to further progress towards an artificial intelligence that can communicate with human users in the form of stories.

## 7. Acknowledgements

We gratefully acknowledge Jill F. Lehman for helping with organizing the annotation effort, Susi Burger and Kory Mathewson for valuable discussions.

## 8. Bibliographical References